\documentclass[10pt,twocolumn,letterpaper]{article}

\usepackage{wacv}
\usepackage{times}
\usepackage{epsfig}
\usepackage{graphicx}
\usepackage{amsmath}
\usepackage{amssymb}
\usepackage{times}
\usepackage{epsfig}
\usepackage{graphicx}
\usepackage{amsmath}
\usepackage{amssymb}
\usepackage{commath}
\usepackage{caption}
\usepackage{wrapfig}
\usepackage{booktabs}       
\usepackage{amsfonts}       
\usepackage{nicefrac}       
\usepackage{microtype}      
\usepackage{mathtools}
\usepackage{cuted}
\usepackage{caption}
\usepackage{subcaption}
\usepackage{breqn}
\usepackage{sidecap}
\usepackage{float}
\usepackage[pagebackref=true,breaklinks=true,letterpaper=true,colorlinks,bookmarks=false]{hyperref}

\wacvfinalcopy 

\def\wacvPaperID{****} 

\ifwacvfinal\fi
\newcommand{\venue}[1]{{\scriptsize #1}}
\begin{document}

\title{\vspace{-5mm}Blended Convolution and Synthesis for Efficient Discrimination of 3D Shapes}

\author{\large Sameera Ramasinghe$^{1,2}$, Salman Khan$^{1,3}$, Nick Barnes$^{1}$ and Stephen Gould$^{1}$\\
\large $^1$Australian National  University, $^2$Data61-CSIRO, $^3$Inception Institute of Artificial Intelligence\\
{\tt\small firstname.lastname@anu.edu.au}}


\maketitle
\ifwacvfinal\thispagestyle{empty}\fi

\begin{abstract}
Existing models for shape analysis directly learn feature representations on 3D point clouds. We argue that 3D point clouds are highly redundant and hold irregular (permutation-invariant) structure, which makes it difficult to achieve inter-class discrimination efficiently. In this paper, we propose a two-pronged solution to this problem that is seamlessly integrated in a single \emph{blended convolution and synthesis} layer. This fully differentiable layer performs two critical tasks in succession. In the \textbf{first} step, it projects the input 3D point clouds into a latent 3D space to synthesize a highly compact and inter-class discriminative point cloud representation. Since, 3D point clouds do not follow a Euclidean topology, standard 2/3D convolutional neural networks offer limited representation capability. Therefore, in the \textbf{second} step, we propose a novel 3D convolution operator functioning inside the unit ball to extract useful volumetric features. We derive formulae to achieve both translation and rotation of our novel convolution kernels. Finally, using the proposed techniques we present an extremely light-weight, end-to-end architecture that achieves compelling results on 3D shape recognition and retrieval.
\end{abstract}


\section{Introduction}

The human world is three-dimensional, therefore optimally understanding and interpreting 3D data is an important research problem. Although deep convolutional neural networks (CNNs) have  been greatly successful in 2D representation learning, they still do not provide an adequate solution to unique challenges that 3D data presents. Specifically, there are two main issues pertinent to 3D data: \emph{(a)} 3D point clouds and rasterized voxel based representations encode redundant information thereby making inter-class discrimination difficult, \emph{(b)} 3D convolutions generally operate in Euclidean space, whereas real-world 3D data lie on a non-Euclidean manifold. The representations thus learned fail to encode the true geometric structure of input shapes. The availability of low-cost 3D sensors and their vast applications in autonomous cars, medical imaging and scene understanding demands a fresh look towards solving the above-mentioned challenges. 

\begin{figure}[t!]
\centering
\includegraphics[width=0.49\textwidth]{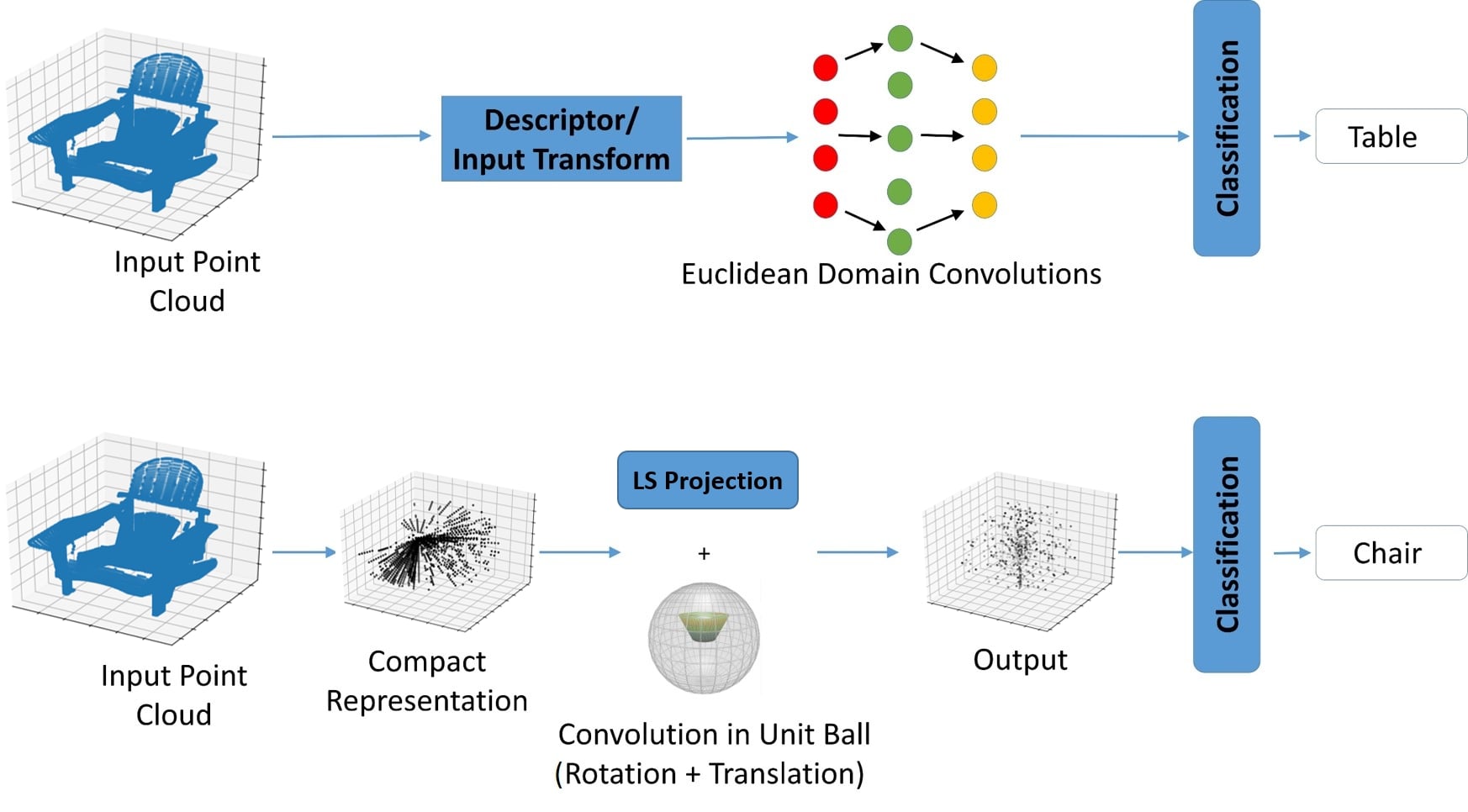}
\caption{High-level comparison of our approach (\emph{bottom)} with the traditional approaches \cite{qi2017pointnet, su2015multi, qi2017pointnet++, klokov2017escape, li2018so} (\emph{top}). We transform an input shape into a compact representation and project it onto a discriminative latent space to capture more discriminative features, before performing convolution in $\mathbb{B}^3$ with roto-translational kernels. Our novel convolution operator has a clear advantage over existing works that only work with Euclidean geometries. This results in a light-weight and highly efficient network design with significantly lower number of layers. }
\label{fig:volsp}
\end{figure}

Existing representation learning schemes for 3D shape description either operate on voxels \cite{brock2016generative, wu2016learning} or point clouds \cite{qi2017pointnet++, klokov2017escape, cheraghian2020, cheraghian2019zeroshot, 8658405, cheraghian2019mitigating, cheraghian2020}. The voxelized data representations are highly sparse, thus prohibiting the design of large-scale deep CNNs.  Efficient data structures such as Octree \cite{meagher1982geometric} and Kdtree \cite{bentley1975multidimensional} have been proposed to solve this problem, however neural networks based representation learning on these tree-based indexing structures is an open research problem \cite{riegler2017octnet}. 
In comparison, point clouds offer an elegant, simple and compact representation for each point $(x,y,z)$. Additionally, they can be directly acquired from the 3D sensors, e.g., low-cost structured light cameras.  On the down side, their irregular structure and high point redundancy pose a serious challenge for feature learning.

We note that recent attempts on direct feature learning from point clouds assume a simplistic pipeline (see Fig.~\ref{fig:volsp}) that mainly aims to extract better global features considering all points \cite{qi2017pointnet, qi2017pointnet++, klokov2017escape, li2018so}. However, all these approaches lack the capacity to work on non-Euclidean geometries and have no inherent mechanism to deal with the high redundancy of point clouds. In this work, we propose an integrated solution, called Blended Convolution and Synthesis (BCS), to address the above-mentioned problems. BCS can effectively deal with the irregular, permutation-invariant and redundant structure of point clouds. Our solution has two key aspects. \emph{First,} we map the input 3D shape into a more discriminative 3D space. We posit that raw 3D point clouds are sub-optimal to be directly used as input to classification models, due to redundant information. This property hampers the classification and retrieval performance by adding an extra overhead to the network, as the network should then disregard redundant features purely using convolution. In contrast, we initially synthesize a more discriminative shape by projecting the original shape to a latent space using a newly derived set of functions which are complete in the unit ball ($\mathbb{B}^3$). The structure of this latent shape is governed by the loss function, and therefore, is optimized to pick up the most discriminative features. This step reduces the number of convolution layers significantly, as shown experimentally in Sec. \ref{sec:experiments}. \emph{Second,} we propose a new convolution operation that works on non-Euclidean typologies i.e., inside the unit ball ($\mathbb{B}^3$). We derive a novel set of complete functions within $\mathbb{B}^3$ that perform convolution in the spectral domain.

Furthermore, since our network operates on the `\emph{spectral domain}', it provides multiple advantages compared to competing models that operate in Euclidean domains: 1) A highly compact and structured representation of 3D objects, which addresses the problem of redundancy and irregularity. Effectively, a 3D shape is represented as a linear combination of complete-orthogonal functions, which allows only a few coefficients to encode shape information, compared to spatial domain representations.  2) Convolution is effectively reduced to a multiplication-like operator which improves computational efficiency, thereby significantly reducing the number of FLOPS. 3) A theoretically sound  way to treat non-Euclidean geometries, which enables the convolution to achieve translational and rotational \emph{equivariance}; and 4) Scalability to large-sized shapes with bounded complexity.

Most importantly, existing methods which perform convolution in the spectral domain \cite{cohen2018spherical, esteves2017learning, ramasinghe2019volumetric} use spherical harmonics or Zernike polynomials to project 3D functions to the spectral domain for performing convolution. The aforementioned function spaces entail certain limitations, e.g.: 1) `Spherical harmonics' only operate on the surface of the unit sphere, which causes critical {information loss} for non-polar shapes. 2) `Zernike polynomials' cause the convolution to achieve only 3D rotational movement of the kernel. In contrast, our newly derived polynomials can handle non-polar shapes, while achieving both 3D rotational and translational movements of the convolution kernel as theoretically proved in Sec.~\ref{sec:convolution}. Recently, Jiang \textit{et al.} \cite{jiang2019convolutional} proposed a novel Fourier transform mechanism to optimally sample non-uniform data signals defined on different topologies to spectral domain without spatial sampling error. This allows CNNs to analyze signals on mixed topologies, regardless of the architecture. However, their spectral transformation does not specifically focus on computational efficiency and equivariance properties, as ours. On the other hand, Jingwei \textit{et al.} \cite{huang2019texturenet} proposed a model which can directly segment textured 3D meshes, by extracting features from high-resolution signals on geodesic neighborhoods of surfaces.
In  contrast, our model consumes point clouds and we propose a lightweight convolution operator, which extracts useful features for 3D classification.

The main contributions of this work are:
\begin{itemize}\setlength{\itemsep}{0em}\vspace{-0.5em}
    \item A novel  approach to obtain a learned 3D shape descriptor, which enhances the convolutional feature extraction process, by projecting the input 3D shape into a latent space, using newly derived functions in $\mathbb{B}^3$.
    \item Develop the theory of a novel convolution operation, which allows both 3D rotational and 3D translational movements of the kernel.
    \item Derive formulae to perform discriminative  latent space projection of the input shape and 3D convolution in a single step, thereby making our approach computationally efficient.
    \item Implement the proposed latent space projection and convolution as a fully differentiable module which can be integrated into any end-to-end learning architecture, and developing a shallow experimental network which produces results on par with state-of-the-art while being computationally efficient.
\end{itemize}

\section{Related Work}
\textbf{3D shape descriptors}: A 3D shape descriptor is a representation of the structural essence of a 3D shape. A variety of hand-crafted feature descriptors have been proposed in past research efforts. A few key such works are based on light field descriptors \cite{chen2003visual}, Fourier transformation \cite{vranic2001tools}, eigen value descriptors \cite{jain2007spectral}, and geometric moments \cite{elad2002content}. Most recent hand-crafted 3D descriptors are based on diffusion parameters \cite{bronstein2010gromov,rustamov2007laplace, bronstein2009shape}. On the other hand, learned 3D shape descriptors have also been popular in the computer vision literature. Litman \etal \cite{litman2014supervised}  propose a supervised bag-of-features (BOF) method to learn a descriptor. Zhu \etal follow an interesting approach, where they first project the 3D shapes into multiple 2D shapes, and then perform training on the 2D shapes to learn a descriptor. Xie \etal \cite{xie2016learned} present a hybrid approach which combines both hand-crafted features and deep networks. They first compute a geometric feature vector from the 3D shape, and then employ a deep network on the feature vector to learn a 3D descriptor. Xie \etal \cite{xie2015deepshape} follow a similar approach, where they first calculate heat kernel signatures of 3D shapes and then use two deep encoders to obtain descriptors. Our work is partially similar to this, but has a key difference: instead of computing hand-crafted features as the first step, we do a learned mapping of input 3D shape into a more discriminative 3D space, which allows us to get rid of high intra-class variances exhibited by most 3D shape descriptors. This step provides another advantage since it maximizes the distance between initial shapes, before being fed to convolution layers later.

\textbf{Orthogonal Moments and 3D Convolution}: Generally, orthogonal moments are used to obtain deformation invariant descriptors from structured data. Compared to geometric moments, orthogonal moments are robust to certain deformations such as rotation, translation and scaling. This property of orthogonal moments has been exploited specially in 2D data analysis in the past \cite{hu1962visual,lin1987classification,arbter1990application,tieng1995application,khalil2001dyadic,suk1996vertex}. Extension of deformation invariant moments from 2D to 3D also has been explored by many prior works \cite{guo1993three,reiss1992features,canterakis19993d,flusser2003moment}. However, the certain properties of these moments depend on the Hilbert space {on which} they are defined. For example, orthogonal moments defined in a sphere or a ball exhibit convenient properties to extract rotation invariants, compared to orthogonal moments defined in a cube. These unique properties of orthogonal moments have recently been used to derive convolution operations which allows 3D rotational movements of kernels \cite{cohen2018spherical,esteves2017learning,ramasinghe2019volumetric, ramasinghe2019representation}. However, the moments used in these works do not contain the necessary properties to achieve 3D translation of the kernels, and therefore, we derive a novel set of functions in $\mathbb{B}^3$ to overcome this limitation. 

\textbf{3D Shape Classification and Retrieval}: Recent works developed for 3D shape classification and retrieval can be broadly categorized into three classes: 1) hand-crafted feature based \cite{vranic2002description}, \cite{guo2016comprehensive} 2) unsupervised learning based \cite{wu2016learning}, \cite{khan2018adversarial} 3) deep learning based \cite{qi2017pointnet,qi2017pointnet++,li2016fpnn}. Generally, deep learning based approaches have shown superior results compared to other two categories. However, the aforementioned deep learning architectures operate on Euclidean spaces, which is sub-optimal for 3D shape analysis tasks, although Weiler \etal \cite{NIPS2018_8239} has shown impressive results using SE(3)-equivariant convolutions in the Euclidean domain. In contrast, our network performs convolution on $\mathbb{B}^3$ which allows efficient feature extraction, since 3D rotation and translation of kernels are easier to achieve in this space.

\section{Preliminaries}
We first provide an overview of basic concepts that will be used later in proposed method. 
\subsection{Complete Orthogonal Systems }

Orthogonal functions are useful tools in shape analysis. Let $\Phi_m$ and $\Phi_n$ be two functions defined in some space $\mathbb{S}$. Then, $\Phi_m$ and $\Phi_n$ are orthogonal over the space $\mathbb{S}$ if and only if,
\begin{equation}
    \int_{\mathbb{S}} \Phi_n(X) \Phi_m(X)dX = 0,\; \forall n \neq m.
\end{equation}
Let $f$ be a function defined in space $\mathbb{S}$, and $\{\Phi_m : m\in \mathbb{Z}^+\}$ be a set of orthogonal functions defined in the same space. Then, the set of orthogonal moments of $f$, with respect to set $\{\Phi_m\}$, can be obtained by $\hat{f}_{m} =  \int_{\mathbb{S}}   f(X) \Phi_m(X)^{\dagger}$
%
where $\dagger$ denotes the complex conjugate. If a set of functions $\{\Phi_m: m\in \mathbb{Z}^+\}$ is both complete and orthogonal, it can reconstruct $f(X)$ using its orthogonal moments as follows,
\begin{equation}
    f(X) = \sum_{m}\hat{f}_{m}\Phi_m(X).
\end{equation}

\subsection{Convolution in Unit Ball $\mathbb{B}^3$}

The unit ball ($\mathbb{B}^3$) is the set of points $x \in \mathbb{R}^3$, where ${\parallel }x {\parallel} {<} 1$. Any point in $\mathbb{B}^3$ can be parameterized using coordinates $(\theta, \phi, r)$, where $\theta, \phi$, and $r$ are azimuth angle, polar angle, and radius respectively. Performing convolution on 3D shapes  in non-linear topological spaces such as the unit ball ($\mathbb{B}^3$) has a key advantage: compared to the Cartesian coordinate system, it is efficient to formulate 3D rotational movements of the convolutional kernel in $\mathbb{B}^3$ \cite{ramasinghe2019volumetric}. To this end, both the input 3D shape and the 3D kernel should be represented as functions in $\mathbb{B}^3$. However, performing convolution in the spatial domain is difficult due to non-linearity of $\mathbb{B}^3$ space \cite{ramasinghe2019volumetric}. Therefore, it is necessary to first obtain the spectral representation of the 3D shape and the 3D kernel, with respect to a set of orthogonal and complete functions in $\mathbb{B}^3$, and consequently perform spectral domain convolution.

\section{Methodology}
Here, we present our `Blended Convolution and Synthesis' layer in detail. First, we construct a set of orthogonal and complete polynomials in $\mathbb{B}^3$. Then, we relax the orthogonality condition of these polynomials, which allows us to project the input shape to a latent space. This projection is a learned process and depends on the softmax cross-entropy between predicted and ground-truth object classes. Therefore, the projected shape is optimized to contain more discriminative properties across object classes. Afterwards, we convolve the latent space shape with roto-translational kernels in $\mathbb{B}^3$ to map it to the corresponding class. Besides, we derive formulae to achieve both projection and convolution in a single step, which makes our approach more efficient.

Below in Section \ref{sec:projection}, we explain the learned projection of the object onto a latent space. Then, in Section \ref{sec:convolution}, we derive our convolution operation, which is able to capture features efficiently using roto-translational kernels.

\subsection{Learned Mapping for Shape Synthesis} 
\label{sec:projection}
In this section, we explain the projection of 3D point clouds to a discriminative latent space in $\mathbb{B}^3$. First, we derive a set of complete orthogonal functions in $\mathbb{B}^3$. Orthogonal moments obtained using orthogonal functions can be used to reconstruct the original object. However, our requirement here is not to reconstruct the original object, but to map it to a more discriminative shape. Therefore, after deriving the orthogonal functions, we relax the orthogonality condition to facilitate the latent space projection. Furthermore, instead of the input point cloud, we use a compact representation as the input to the feature extraction layer, for efficiency and to leverage the capacity of convolution in $\mathbb{B}^3$. In Section \ref{sec:compact}, we explain our compact representation.

\subsubsection{Compact Representation of Point Clouds}
\label{sec:compact}
Most 3D object datasets contain point clouds with uniform texture. That is, if the 3D shape is formulated as a function $f$ in $\mathbb{B}^3$, such that for any point on the shape, $f(\theta, \phi, r) = c$, where $c$ is a constant. However, formulating 3D shapes in $\mathbb{B}^3$ has the added advantage of representing both 2D texture and 3D shape information simultaneously \cite{ramasinghe2019volumetric}. Therefore, the advantage of convolution in $\mathbb{B}^3$ can be utilized when the input and kernel functions have texture information.

Following this motivation, we convert the uniform textured point clouds into non-uniform textured point clouds using the following approach. First, we create a grid using equal intervals along $r,\theta$, and $\phi$. We use $25, 36$, and $18$ interval spaces for $r,\theta$, and $\phi$, respectively. Then, we bin the point cloud to grid points, which results in a less dense, non-uniform surface valued point cloud. The obtained compact representation does not contain all the fine-details of the input point cloud. However in practice, it allows better feature extraction using the kernels. A possible reason could be that kernels are also non-uniform textured point clouds with discontinuous space representations, and they can capture better features from non-uniform textured input point clouds when performing convolution in $\mathbb{B}^3$.

\subsubsection{Derivation of orthogonal functions in $\mathbb{B}^3$}
\label{sec:orthogonal}
In this section, we derive a novel set of orthogonal polynomials with necessary properties to achieve the translation and rotation of convolution kernels. Afterwards, in Section \ref{sec:relaxation}, we relax the orthogonality condition of the polynomials to facilitate latent space projection.

Canterakis \etal \cite{canterakis19993d} showed that a set of orthogonal functions which are complete in unit ball can take the form $Z_{n,l,m}(r,\theta, \phi) = Q_{nl}(r)Y_{l,m}(\theta, \phi)$,
where $Q_{nl}$ is the linear component and $Y_{l,m}(\theta, \phi)$ is the angular component. The variables $r$, $\theta$ and $\phi$ are radius, azimuth angle and polar angle, respectively. We choose $Y_{l,m}(\theta, \phi)$ to be spherical harmonics, since they are complete and orthogonal in $S^2$.

For the linear component, we do not use the Zernike linear polynomials as in Canterakis \etal \cite{canterakis19993d}, as they do not contain the necessary properties to achieve the translational behaviour of convolution kernels \cite{ramasinghe2019volumetric}. Therefore, we derive a novel set of orthogonal functions, which are complete in $0 < r < 1$, and can approximate any function in the same range. Furthermore, it is crucial that these functions contain necessary properties to achieve the translation of kernels while performing convolution. Therefore, we choose the following function as the base function:
\begin{equation}
    \label{equ:f}
    f_{nl} = (-1)^ln \sum_{k=0}^{n}\frac{((n-l)r)^k}{k!}.
\end{equation}
It can be seen that,  
\begin{equation}\label{eq:approx}
  f_{nl}  \approx  (-1)^ln\exp(r(n-l)),
\end{equation}
as $n$ increases, for small $r$. Therefore, we use the approximation given in Eq.~\ref{eq:approx} in future derivations. As we show in Section \ref{sec:convolution}, this property is vital for achieving the translation of kernels. Next, we orthogonalize $f_{nl}(r)$ to obtain a new set of functions $Q_{nl}(r)$. Consider the orthogonality $\int_{\mathbb{B}^3} Z_{n,l,m}Z_{n',l',m'} = 0, \forall n \neq n', l\neq l', m \neq m'$. If we consider only the linear component, {the} orthogonality condition should be $\int_{0}^{1} Q_{n,l} Q_{n',l'}r^2dr = 0, \forall n \neq n', l \neq l'$. Therefore, $Q_{n,l}$ should be orthogonal with respect to the weight function $w(r) = r^2$. We define,
\begin{equation}
    \label{equ:q}
    Q_{nl}(r) = f_{nl}(r) - \sum_{k=0}^{n-1} \sum_{m=0}^{k}C_{nlkm}Q_{km}(r)
\end{equation}
where $n \geq 0, n \geq l \geq 0$ and $C_{nlkm}$ is a constant. Since $Q_{nl}$ should be an orthogonal set, the inner product between any two different $Q_{nl}$ functions is zero. Therefore, we obtain,
\begin{align}
    \langle Q_{nl},Q_{n'l'} \rangle & = \langle f_{nl}, Q_{n'l'} \rangle   - \sum_{k=0}^{n-1} \sum_{m=0}^{k}C_{nlkm}\langle Q_{km}, Q_{n'l'}\rangle \notag 
\end{align}
Since $\langle Q_{nl},Q_{n'l'} \rangle =0$, we get:
\begin{equation}
    \label{equ:c}
    C_{nln'l'} = \frac{\langle f_{nl}, Q_{n'l'}\rangle }{\parallel Q_{n'l'}\parallel^2}.
\end{equation}
Following this process, we can obtain the set of orthogonal functions $Q_{nl}$ for $n \geq 0, n \geq l$. The derived polynomials up to $n=5,l=5$ are shown in Appendix A. In Section \ref{sec:completeness}, we prove the completeness property of the derived functions.

\subsubsection{Completeness in $\mathbb{B}^3$}
\label{sec:completeness}
In this section, we prove the completeness in $\mathbb{B}^3$ for the set of functions $\{Q_{nl}\}$ derived in Section \ref{sec:orthogonal}.

\noindent \textbf{Condition 1}:\textit{ Consider the orthogonal set $\{p_n\}$ defined in $L^2[0,1]$. Then, $\{p_n\}$ is complete in space $L^2[0,1]$ if and only if there is no non-zero element in $L^2[0,1]$ that is orthogonal to every $\{p_n\}$}. 

To show that $f_{nl}$ is complete over $L^2[0,1]$, we first prove the completeness of the set $\{\Phi_n\}$, which is obtained by orthogonalizing the set $\{1,x,x^2,x^3,...\}$. Let $\Psi(x)$ be an element in $L^2[0,1]$, which is orthogonal to every element of $\{1,x,x^2,x^3,...\}$. Then, suppose the following relationship is true:
\begin{equation}
\label{equ:fourier}
    \langle \Psi, e^{2\pi ikx} \rangle = \sum_{n=0}^\infty \frac{(2 \pi ikn)^n}{n!} \langle \Psi, x^n \rangle = 0,
\end{equation}
where $k$ is a constant. However, we know that $\{e^{2\pi ikx}\}_{k=0}^{k=\infty}$ is the complex exponential Fourier basis, and is both complete and orthogonal. 
Therefore, if Eq.~\ref{equ:fourier} is true, $\Psi = 0$, which gives us the result, i.e., $\langle \Psi, x^n \rangle = 0 \equiv \,\Psi=0$. Equivalently, since $\{\Phi_n\}$ is obtained by orthogonalization of $\{1,x,x^2,x^3,...\}$, $\langle \Psi, \{\Phi_n\} \rangle = 0 \equiv\, \Psi$ = 0. Hence, according to Condition 1, $\{\Phi_n\}$ is complete in $L^2[0,1]$.

Next, we consider the set $Q_{n,l}$. Since $Q_{n,l}$ is orthogonalized using the basis functions in Eq.~\ref{equ:f}, 
it is enough to show that $f_{nl}$ is complete over $L^2[0,1]$. Let $\Theta$ be a function defined in $L^2[0,1]$. Then, suppose the following relationship is true:
\begin{equation}
    \label{equ:complete}
    \langle \Theta, f_{n,l}\rangle = (-1)^ln \sum_{k=0}^{n}\frac{((n-l)^k}{k!} \langle \Theta , r^k \rangle = 0.
\end{equation}
For Eq. \ref{equ:complete} to be true,  $\langle \Theta , r^k \rangle = 0$ for $k=\{0,1,2,...\}$. But we showed that this condition is satisfied if and only if $\Theta = 0$. Therefore, $\langle \Theta, f_{n,l}\rangle = 0, \forall n \geq l \geq 0  \equiv \,\Theta = 0$. Hence, $f_{n,l}$ is complete in $L^2[0,1]$.

\subsubsection{Relaxation of orthogonality of functions in $\mathbb{B}^3$}
\label{sec:relaxation}
Computing $C_{nln'l'}$ using Eq. \ref{equ:c} ensures the orthogonality of $Q_{nl}$. Since $Q_{nl}$ and $Y_{lm}$ are both orthogonal and complete, projecting the input shape $f$ onto the set of functions $Z_{nlm}, n \geq l \geq m \geq 0$, enables us to reconstruct $f$ by:
\begin{equation}
\label{equ:reconstruction}
f(\theta, \phi, r) = \sum\limits_{n=0}^{\infty} \sum\limits_{l = 0}^{n} \sum\limits_{m = -l}^{l} \Omega_{n,l,m}(f) Z_{n,l,m}(\theta, \phi, r),
\end{equation}
where spectral moment $\Omega_{n,l,m}(f)$ can be obtained using $\Omega_{n,l,m}(f) = \int_{0}^{1}\int_{0}^{2 \pi}\int_{0}^{\pi}  f(\theta, \phi, r) {Z}_{n,l,m} r^2 \sin\phi \, drd\phi d\theta.\notag$.
Representing $f$ in spectral terms, as in Eq.~\ref{equ:reconstruction}, enables easier convolution in spectral space, as derived in Section \ref{sec:convolution}.

\begin{figure*}[!htp]
\centering
\includegraphics[width = 1.0\textwidth]{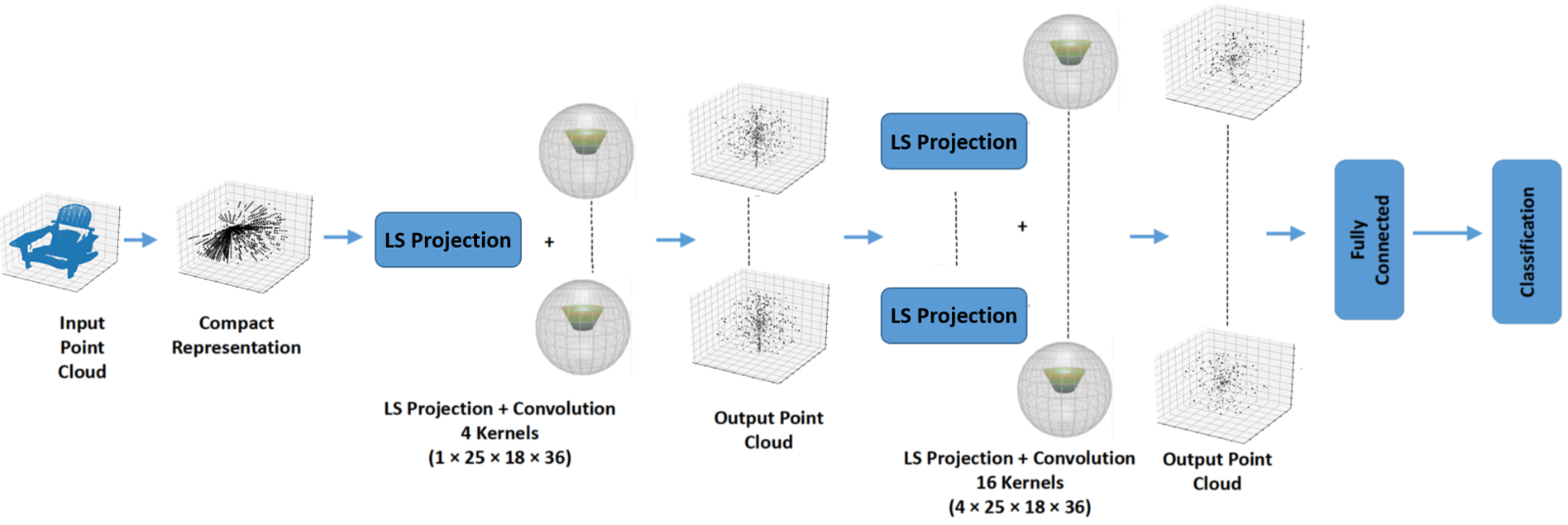}
\caption{The overall CNN architecture. Our proposed design is a light-weight model, comprising of only three weight layers. Our networks aims to achieve a compact latent representation and volumetric feature learning via convolutions in $\mathbb{B}^3$. }
\label{fig:archi}
\end{figure*}

However, we argue that since 3D point clouds across different object classes contain redundant information, projecting the point clouds in to a more discriminative latent space can improve classification accuracy. Our aim here is to reduce redundant information and noise from the input point clouds and map it to a more discriminative point cloud, which concentrates on discriminative geometric features. Therefore, we make $C_{nln'l'}$ trainable, which allows the latent space projection $\hat{f}$ of the input shape $f$ as follows:
\begin{equation}
\label{equ:reconstruction_latent}
\hat{f}(\theta, \phi, r) = \sum\limits_{n=0}^{\infty} \sum\limits_{l = 0}^{n} \sum\limits_{m = -l}^{l} \hat{\Omega}_{n,l,m}(f) \hat{Z}_{n,l,m}(\theta, \phi, r),
\end{equation}
where spectral moment $\hat{\Omega}_{n,l,m}(f)$ can be obtained using,
\begin{equation}
\hat{\Omega}_{n,l,m}(f) = \int_{0}^{1} \int_{0}^{2 \pi} \int_{0}^{\pi}  f(\theta, \phi, r) {\hat{Z}}^{\dagger}_{n,l,m} r^2 \sin\phi \, drd\phi d\theta, \notag
\end{equation}
where $\hat{Z}_{n,l,m}(\theta, \phi, r) = \hat{Q}_{nl}(r)Y_{lm}(\theta, \phi)$ and $\hat{Q}_{nl}(r) = f_{nl}(r) - \sum_{k=0}^{n-1} \sum_{m=0}^{k}W_{nlkm}\hat{Q}_{km}(r)$.
Here, the set $\{W_{nlkm}\}$ denotes trainable weights. Note that since the final orthogonal function is a product of the linear and the angular parts, making both functions learnable is redundant.

\subsection{Convolution of functions in $\mathbb{B}^3$}
\label{sec:convolution}
Let the north pole be the $y$ axis of the Cartesian coordinate system and the kernel is symmetric around $y$. Let $f(\theta, \phi, r)$, $g(\theta, \phi, r)$ be the functions of object and kernel respectively. Then, convolution of functions in $\mathbb{B}^3$ is defined by:
\begin{align}
\label{conveq}
& f * g(\alpha, \beta,r') \coloneqq  \langle f(\theta, \phi, r), T_r'\{\tau_{(\alpha, \beta)}(g(\theta, \phi, r))\}\rangle \\ 
& {=}{\int_{0}^1}{\int_{0}^{2\pi}}{\int_{0}^{\pi}}f(\theta, \phi, r)T_r'\{\tau_{(\alpha, \beta)}(g(\theta, \phi, r))\}\sin\phi \, d\phi d\theta dr, \notag
\end{align}
where $\tau_{(\alpha, \beta)}$ is an arbitrary rotation that aligns the north pole with the axis towards the $(\alpha, \beta)$ direction ($\alpha$ and $\beta$ are azimuth and polar angles respectively) and $T_r'$ is translation by $r'$.






To achieve both latent space projection and convolution in $\mathbb{B}^3$ in single step, we present the following theorem.

\noindent
\textbf{Theorem 1: } \textit{Suppose $f,g : X \longrightarrow \mathbb{R}^{3}$ are square integrable functions defined in $\mathbb{B}^{3}$ so that $\langle f,f \rangle < \infty$ and $\langle g,g \rangle < \infty$. Further, suppose $g$ is symmetric around {the} north pole and $\tau (\alpha, \beta) = R_y(\alpha)R_z(\beta)$ where $R \in \mathbb{SO}(3)$ and $T_r'$ is translation of each point by $r'$. Then, }
\begin{align}\small
    & \int_{0}^1\int_{0}^{2\pi}\int_{0}^{\pi}P\{f(\theta, \phi, r)\} T_r'\{\tau_{(\alpha, \beta)}(g(\theta, \phi, r))\}\sin\phi\, d\phi d\theta dr \notag\\
    & \approx  \frac{4 \pi}{3} \sum\limits_{n=0}^{\infty} \sum\limits_{l = 0}^{n} \sum\limits_{m = -l}^{l}  \langle f_{nl}(r),  Q_{n'l}(r) \rangle ( e^{(n-l)r'}-  e^{(n'-l)r'}) \notag\\
    & \qquad \hat{\Omega}_{n,l,m}(f) \hat{\Omega}_{n,l,0} (g) Y_{l,m}(\theta, \phi), 
\end{align}
where, $\hat{\Omega}_{n,l,m}(f), \hat{\Omega}_{n,l,0}(g)$ and $Y_{l,m}(\theta, \phi)$ are $(n,l,m)^{th}$ spectral moment of $f$, $(n,l,0)^{th}$ spectral moment of $g$, and spherical harmonics function, respectively. $P\{\cdot\}$ is the projection to a latent space, $\tau (\alpha, \beta) = R_y(\alpha)R_z(\beta)$ where $R \in \mathbb{SO}(3)$ and $T_r$ is translation of each point by $r$. The proof to this theorem can be found in Appendix A.

\subsection{Network Architecture} 
Our experimental architecture consists of two convolution layers and a fully connected layer. We employ four kernels in the first convolution layer and 16 kernels in the second convolution layer, each followed by group normalization \cite{wu2018group} and a ReLU layer. The experimental architecture is illustrated in Figure \ref{fig:archi}. We use $n=5$ for implementing Eq.~\ref{conveq} and softmax cross-entropy loss as the objective function during training. For training, we use a two step process. First, we train polynomial weights using a learning rate of $10^{-5}$, and then train kernel weights using a learning rate of $0.01$. We used {the} Adam optimizer for calculating gradients with parameters $\beta_1 = 0.9$, $\beta_2 = 0.999$, and $\epsilon = 1 \times 10^{-8}$, where parameters refer to {the} usual notation. We use $20k$ iterations to train polynomials weights and $30k$ iterations to train kernel weights. We use a single GTX 1080Ti GPU for training and the model takes around 30 minutes to complete a single epoch during training on ModelNet10 dataset.

\begin{table*}[!htp]
\scriptsize
  \begin{minipage}[b]{0.70\hsize}\centering
  \begin{tabular}{lccccc}
    \toprule
    \textbf{Method}     & \textbf{Modality} & \textbf{Views} & \#\textbf{Layers} &   \textbf{ModelNet10}   & \textbf{ModelNet40}  \\
    \midrule
  
    VoxNet \venue{(IROS'15)} \cite{maturana2015voxnet} & Volume & - & - &  92.0\% & 83.0\%\\
    
    3DGAN \venue{(NIPS'16)} \cite{wu2016learning} & Volume &  - & - & 91.0\% & 83.3\%\\
    
    3DShapeNet \venue{(CVPR'15) } \cite{wu20153d}  & Volume &     & 4-3DConv + 2FC & 83.5\% & 77\%\\
    
             VRN (\venue{(NIPS'16)}) \cite{brock2016generative} & Volume & - & 45Conv &  93.6\%   & 91.3 \% \\
    \midrule
    
    GIFT (\venue{(CVPR'16)}) \cite{bai2016gift} & RGB & 64 & - &  92.4\% & 83.1\% \\
    
            Pairwise (\venue{(CVPR'16)}) \cite{johns2016pairwise} & RGB & 12 &    23Conv &  92.8\% & 90.7\%      \\
      MVCNN \venue{(ICCV'16)} \cite{su2015multi}  & RGB & 12   & 60Conv + 36FC & - & 90.1\%     \\
      MHBN \venue{(CVPR'18)} \cite{yu2018multi} & RGB & 6& 78Conv + 18FC & 95.0\% & \textbf{94.7\%}\\
       
                      DeepPano \venue{(SPL'15)} \cite{shi2015deeppano} & RGB &    & 4Conv + 3FC & 85.5\% & 77.63\%     \\
    \midrule
    ECC \venue{(CVPR'17)} \cite{simonovsky2017dynamic} & Points & -    & 4Conv + 1FC &  90.0\% & 83.2\%    \\
      
          Kd-Networks \venue{(ICCV'17)} \cite{klokov2017escape} & Points & -  &  15KD &  93.5\% &  91.8\%     \\
          
          SO-Net \venue{(CVPR'18)} \cite{li2018so} & Points & & 11FC&  \textbf{95.7\%} & 93.4\%       \\
          
          PointNet \venue{(CVPR'17)} \cite{qi2017pointnet} & Points & -  & 5Conv + 2STL    &   -& 89.2\%      \\
          
          
          {LP-3DCNN \venue{(CVPR'19)} \cite{kumawat2019lp}} & Points & - & 15Conv + 3FC &  - & 92.1\% \\

 Ours &  Points & - & 2Conv + 1FC & 94.2\% & 91.8\%\\




          
              

     
    \bottomrule
  \end{tabular}
  \captionof{table}{Model accuracy vs depth analysis on ModelNet10 and ModelNet40 datasets.}
    \label{table:mod40}
   \end{minipage}
   \hfill
   \begin{minipage}[b]{0.3\hsize}\centering
   \begin{tabular}{lcc}
   \toprule
   \textbf{Method} & \textbf{FLOPS} &  \textbf{ModelNet40} \\ 
     & (inference) & \\
   \midrule
   PointNet \cite{qi2017pointnet} & 14.70B & 89.2\% \\
   SpecGCN \cite{wang2018local} & 17.79B  & 92.1\%\\
   PCNN \cite{atzmon2018point} & 4.70B  & 92.3\%\\ 
   PointNet++ \cite{qi2017pointnet++} & 26.04B & 91.9\%\\
   3DmFV-Net \cite{ben20173d} & 16.89B & 91.6\%\\
   PointCNN \cite{li2018pointcnn} & 25.30B & 92.2\%\\
   DGCNN \cite{wang2018dynamic} & 44.27B & \textbf{93.5\%}\\
   Ours & \textbf{1.31}B  & 91.8\%\\
   \bottomrule
   \end{tabular}
   \caption{  
      Our model complexity is much lower compared to state-of-the-art 3D classification models. The FLOPS (inference time) comparisons are reported according to \cite{li2018pointcnn} settings with 16 batch size.}
  \label{tab:complexity}
   \end{minipage}
\end{table*}

\section{Experiments}
\label{sec:experiments}
We evaluate the proposed methodology on 3D object classification and 3D object retrieval tasks using recent datasets: ModelNet10, ModelNet40, McGill 3D, SHREC’17 and OASIS. We also conduct a thorough ablation study to demonstrate the effectiveness of our derivations and design choices.

\subsection{3D Object Classification Performance}

A key feature of our proposed pipeline is the projection of the input 3D shapes into a more discriminative latent shape, before feeding them into convolution layers. One critical advantage of this step is that original subtle differences across object classes are magnified in order to leverage the feature extraction capacity of convolution layers. Therefore, the proposed network should be able to capture more discriminative features in the lower layers, and provide better classification results with a smaller number of layers, compared to other state-of-the-art works which directly extract features from the original shape. To illustrate this, we present a model depth vs accuracy analysis on ModelNet10 and ModelNet40 in Table \ref{table:mod40}, and compare the effectiveness of our network with other comparable state-of-the-art approaches.

State-of-the-art work can be mainly categorized into three types: volume based, RGB based and Points based. Volume based methods generally rely on volumetric representation of the 3D shape such as voxels. VoxNet \cite{maturana2015voxnet} shows the best performance among volume based models, with an accuracy of 92.0\% in ModelNet10 and 83.0\% in ModelNet40, which is lower than our model's accuracy. It is interesting to see that 3DShapeNets \cite{wu20153d}, and VRN \cite{brock2016generative} have  significantly more layers compared to our model, although accuracies are lower. In general, our model performs better and has a lower model depth compared to volume based methods.

RGB based models generally follow the projection of the 3D shape into 2D representations, as an initial step for feature extraction. We perform better than all the RGB based methods, except for MHBN \cite{yu2018multi}, which has accuracies 95.0\% and 94.7\% over ModelNet10 and ModelNet40 respectively. However, MHBN contains six views and for each view they employ a VGG-M network for initial feature extraction. This results in a significantly complex setup, which contains 96 trainable layers. In contrast, our model uses a single view and three trainable layers. Generally, RGB based models use multiple views, pre-trained deep networks and ensembled models, which results in increased model complexity. In contrast, our model use a single view and does not use pre-trained models, and achieves the second highest performance compared to RGB based models.

Point based models directly consume point clouds. Our model achieves the second best performance in this category, the highest being SO-NET \cite{li2018so}. However, SO-NET contains 11 fully connected layers, while our model only contains three layers. Our model is able to outperform the other point based setups, although their model depths are larger. 

Overall, our model achieves a performance mark comparable to the best models, with a much shallower architecture. Our model contains the lowest number of trainable layers compared to all the models. This analysis on ModelNet10 and ModelNet40 clearly reveals the efficiency and better feature extraction capacity of our approach. Table \ref{tab:complexity} depicts the computational efficiency of BCS compared to state-of-the-art. With just $1.31$B FLOPs, we outperform the closest contender PCNN \cite{atzmon2018point} by a significant $3.39$B margin.

\subsection{3D Object Retrieval Performance}
\begin{figure}
\centering
\begin{subfigure}{.25\textwidth}
  \centering
  \includegraphics[width=1\linewidth]{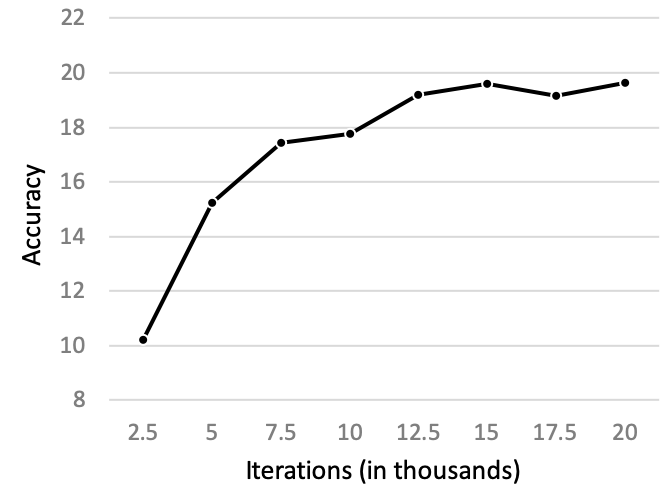}
  \label{fig:sub2}
\end{subfigure}%
\begin{subfigure}{.25\textwidth}
  \centering
  \includegraphics[width=1\linewidth]{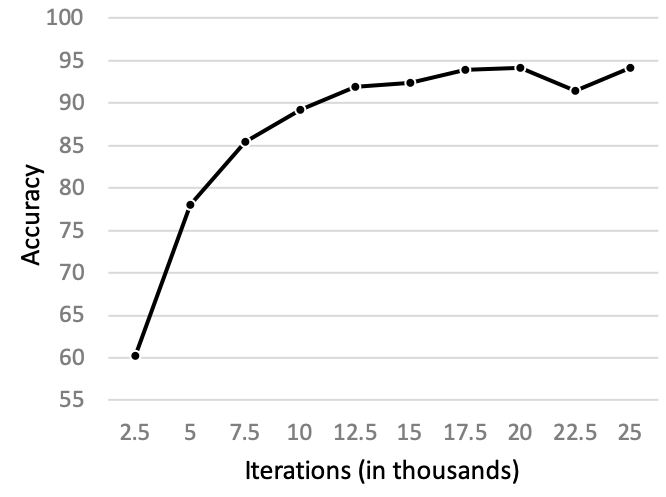}
  \label{fig:sub2}
\end{subfigure}\vspace{-2em}
\caption{Training curves of our architecture on ModelNet10 for polynomial weights (\emph{left}) and kernel weights (\emph{right}).}
\label{fig:trainng}
\end{figure}

 In this section, we compare the performance of our approach in 3D object retrieval. We use the McGill 3D dataset and SHREC'17 dataset for our experiments. We first obtain the feature vectors computed by each kernel in the second layer, and concatenate them. Then, we apply an autoencoder on the concatenated vector and retrieve a 1000-dimensional descriptor. Then we measure the cosine similarity between input and target shapes to measure the 3D object retrieval performance. We use the nearest neighbour performance and the evaluation metric. Table \ref{table:mcgill} depicts the results on the McGill Dataset. Out of the six state-of-the-art models compared, our model achieves the  best retrieval performance. Table \ref{table:shrec} illustrates the performance comparison on the SHREC'17 dataset, where our approach gives the second best performance, below Furuya \etal \cite{Furuya2016DeepAO}. Figure \ref{fig:trainng} depicts our training curves for polynomial weights and kernel weights. The training curves are obtained for ModelNet10.
 
 \subsection{Ablation Study}
\begin{figure}
\centering
\begin{subfigure}{.25\textwidth}
  \centering
  \includegraphics[width=1\linewidth]{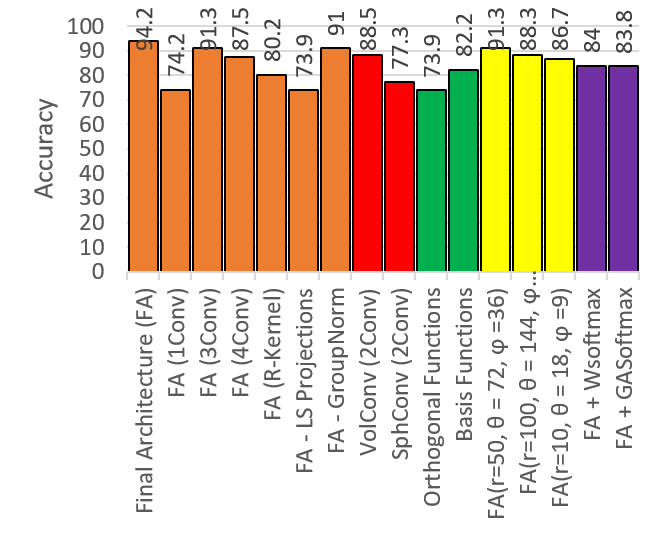}
  \label{fig:sub1}
\end{subfigure}%
\begin{subfigure}{.25\textwidth}
  \centering
  \includegraphics[width=0.8\linewidth]{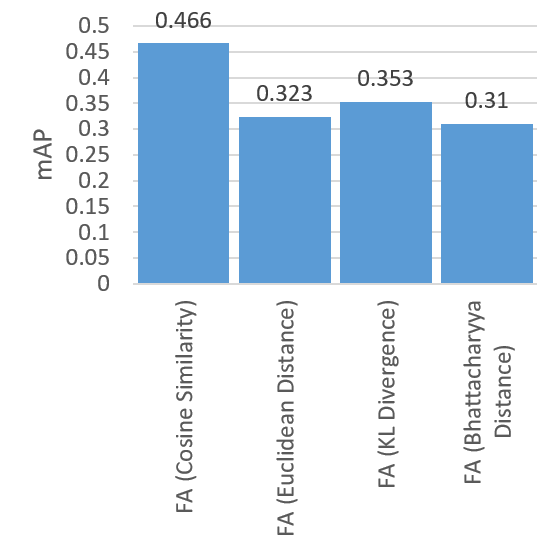}
  \label{fig:sub2}
\end{subfigure}\vspace{-1.5em}
\caption{Ablation study on ModelNet10 in 3D object classification (\emph{left}) and SHREC'17 in 3D object retrieval (\emph{right}).}
\label{fig:ablation}
\end{figure}

 In this section, we conduct an ablation study on our model and discuss various design choices, as illustrated in Figure \ref{fig:ablation}. Firstly, we use a single convolution layer instead of two, and achieve an accuracy of 74.2\% over ModelNet10. Then, we investigate the effect of using a higher number of convolution layers. We get accuracies 91.3\% and 87.5\%, when using three and four convolution layers respectively. Therefore, using two convolution layers yields the best performance. An important feature of our convolution layer is the translation of convolution kernels, in addition to rotation. To evaluate the effect of this, we use only rotating kernels and measure the performance, and achieve an accuracy of 80.2\%. Therefore, it can be concluded that having the translational movements of the kernel has caused an accuracy increment of 14\%, which is significant. Next, we measure the effect of latent space projection. To this end, we use orthogonal polynomials derived in Equations \ref{equ:q}-\ref{equ:c} for convolution, instead of making them learnable. This removes the latent space projection of the input, as the original object is reconstructed using spectral moments. After removing the latent space projection, the accuracy is dropped by 20.3\%, which clearly reveals the significance of this feature. Then, we replace our convolution layers with volumetric convolution \cite{ramasinghe2019volumetric} layers and spherical convolution layers \cite{cohen2018spherical} and get 88.5\% and 77.3\% accuracy respectively. This shows that our convolution layer has a better feature extraction capacity compared to other convolution operations. One key reason behind this can be the translational movements of our kernels and the combined latent space projection step, which the aforementioned convolution methods lack.
 
 Moreover, we test our model using basis functions in Eq.~\ref{equ:f} as the projection functions, instead of learnable functions. Also, we again test the model using orthogonal functions. In both cases, the performance is lower compared to learnable functions. Furthermore, instead of soft-max cross entropy, we use WSoftmax \cite{liu2017deep} and GASoftmax \cite{liu2017deep} and achieve only 84.0\% and 83.0\% respectively. Therefore, using soft-max cross entropy as the loss function is justified. We also evaluate the effect of sampling density on accuracy. As shown in Figure \ref{fig:ablation}, accuracy drops below 94.2\%---which is reported by final architecture---when using a denser representation. Similarly, accuracy drops to 86.7\% when using $r=10, \theta = 18, \phi=9$ as sampling intervals. Therefore, using $r=25, \theta = 36, \phi=18$ as in the final architecture seems to be the ideal design choice. We use four different distance measures in the 3D object retrieval task and compare the performance: cosine similarity, Euclidean distance, KL divergence, and Bhattacharya distance. Out of these, cosine similarity yields the best performance, with a mAP of 0.466.

 
 
 
\begin{SCtable}[][tp]
\scriptsize
  \caption{3D object retrieval results comparison with state-of-the-art on McGill Dataset.}
  \label{table:mcgill}
  \centering
  \begin{tabular}{ll}
    \toprule
    \cmidrule(r){1-2}
    Method     & Accuracy  \\
    \midrule
    Bashiri \etal \cite{bashiri2019application} \venue{(arxiv'19)} & 0.9646\%\\
    
    Zeng \etal \cite{zeng2018dempster} \venue{(IET'18)}  & 0.981\%\\
    Han \etal \cite{han2018deep} \venue{(IP'18)} & 0.8827\% \\
    Tabia \etal \cite{tabia2014covariance} \venue{(CVPR'14)}& 0.977\% \\
    Papadakis \etal \cite{papadakis20083d} \venue{(3DOR'w)} & 0.957\% \\
    Lavoue \etal \cite{lavoue2012combination} \venue{(TVC'14)} & 0.925\% \\
    Xie \etal \cite{xie2015deepshape} \venue{(CVPR'15)} & 0.988\% \\
    \midrule
    \textbf{Ours} & \textbf{0.990\%} \\
               
     
    \bottomrule
  \end{tabular}
\end{SCtable}

\begin{SCtable}[][htp]
\scriptsize
  \caption{3D object retrieval results comparison with state-of-the-art on SHREC'17.}
  \label{table:shrec}
  \centering
  \begin{tabular}{ll}
    \toprule
    \cmidrule(r){1-2}
    Method     & mAP  \\
    \midrule
Furuya \etal \cite{Furuya2016DeepAO} \venue{(BMVC'16)} & \textbf{0.476} \\
Esteves \etal \cite{esteves2017learning} \venue{(ECCV'18)}  & 0.444 \\
Tatsuma \etal \cite{Tatsuma2009} \venue{(TVC'09)} & 0.418 \\
Bai \etal \cite{bai2016gift} \venue{(CVPR'16)} & 0.406 \\
\midrule
Ours & 0.466 \\

    \bottomrule
  \end{tabular}
\end{SCtable}

\subsection{Classification of Complex Shapes}
The proposed convolution layer offers two key advantages: 1) the ability to simultaneously model both shape and texture information, and 2) handling non-polar objects. However, we used ModelNet10 to conduct the ablation study shown in Figure~\ref{fig:trainng}, which contains relatively simple shapes (i.e. not dense in $\mathbb{B}^3$), and it is clear from the results that the accuracy drops when more than two convolution layers are used. A possible reason for this behaviour is overfitting. Since our convolution layer can capture highly discriminative features from the input functions, using more parameters can cause overfitting on relatively simpler shapes, and thus, a drop in classification accuracy. To test this hypothesis, we conduct an experiment on a more challenging dataset, which contains highly non-polar and textured objects.

In this experiment, we use OASIS-3 dataset \cite{fotenos2008brain} to sample $1000$ 3D brain scan images. The dataset includes brain scan images from both Alzheimer's disease patients and healthy subjects. A key property of these images is that they have texture information and are highly dense in $\mathbb{B}^3$. Firstly, we split the sampled data in to train and test sets, with $800$ and $200$ images for each set, respectively. To avoid bias, we include an equal number of Alzheimer cases and healthy cases in both train and test sets. Then, we evaluate different network architectures using the
dataset, varying the number of convolution layers. We use cross entropy loss function in this experiment. The results are shown in Table \ref{tab:multilayer}.

\begin{SCtable}
    \caption {Multi-layer architectures for highly non-polar and textured shape classification. Our model shows an improvement with more layers.}
\centering
\label{tab:multilayer}
\scalebox{0.8}{
 \begin{tabular}{  c  c   }
     \toprule
 \textbf{Model} & \textbf{Accuracy}  \\ 
\hline
Ours (1 Conv layer) & 66.3\% \\ 
Ours (2 Conv layers) & 82.7\%\\
Ours (3 Conv layers) &  86.7\% \\
Ours (4 Conv layers) &  \textbf{88.1\%}\\
Ours (5 Conv layers) &  87.0\%\\
\bottomrule
  \end{tabular}}
   \end{SCtable}
   
As evident from Table \ref{tab:multilayer}, the classification accuracy increases with the number of convolution layers, up to four layers. Hence, it can be concluded that more challenging objects allow our model to demonstrate its full capacity.

\section{Conclusion}
In this paper, we propose a novel approach called `Blended Convolution and Synthesis' to analyse 3D data, which entails two key operations: (1) learning a 3D descriptor obtained by projecting the input 3D shape into a discriminative latent space and (2) convolving the 3D descriptor in $\mathbb{B}^3$ with roto-translational 3D kernels for extracting features. We derive a novel set of polynomials in $\mathbb{B}^3$, and project the input data into a spectral space using the derived polynomials to join these two operations into a single step. Furthermore, we use a compact representation of the input data to reduce the density of the data distribution and leverage the advantage of convolving functions in  $\mathbb{B}^3$. Finally, we present a light-weight architecture and achieve compelling results in 3D object classification and 3D object retrieval tasks.

{\small
\bibliographystyle{ieee}
\bibliography{egbib}
}

\newpage
\appendix
\onecolumn

\begin{center}
{\Large\bf Appendix A\\[0.5em]
Blended Convolution and Synthesis for Efficient Discrimination of 3D Shapes}
\end{center}
\vspace{1em}

\section{The derived $Q_{nl}$ polynomials up to $n{=}5$, $m{=}5$:}
$Q_{nl}$ polynomials up to  $n=5$ and $m=5$ are shown in Table \ref{table:q}.

\begin{table*}[h]
  \caption{The derived $Q_{nl}$ polynomials up to $n=5$, $m=5$.}
  \label{table:q}
  \centering
  \begin{tabular}{ll}
    \toprule
    \cmidrule(r){1-2}
  Polynomial     & Expression   \\
  \midrule
  $Q_{00}$ & $0$ \\
  $Q_{10}$ & $1. + 2x$ \\
  $Q_{11}$ & $-1. - 1x$ \\
  $Q_{20}$ & $-9.79 - 10.65x + 9x^2$ \\
  $Q_{21}$ & $5.29 + 6.29x - 4x^2$ \\
  $Q_{22}$ & $-1.99 - 3.63x + x^2$ \\
  $Q_{30}$ & $-123.58 - 158.11x + 87.46x^2 + 32x^3$ \\
    $Q_{31}$ & $70.26 + 89.41x - 50.31x^2 - 13.5x^3$ \\
    $Q_{32}$ & $15.86 + 22.27x - 11.06x^2 - 0.5x^3$ \\
    $Q_{33}$ & $-768.81 - 1006.25x + 512.65x^2 + 139.10x^3 + 104.16x^4$ \\
    $Q_{40}$ & $-35.86 - 46.15x + 25.59x^2 + 4x^3$ \\
    $Q_{41}$ & $422.87 + 550.70x - 287.81x^2 - 73.52x^3 - 42.66x^4$ \\
    $Q_{42}$ & $-768.81 - 1014.25x + 480.65x^2 + 73.77x^3 + 13.5x^4$ \\
    $Q_{43}$ & $-776.81 - 1034.25x + 454.65x^2 + 50.43x^3 - 2.66x^4$ \\
    $Q_{44}$ & $-768.81 - 1022.25x + 464.65x^2 + 56.43x^3 + 0.16x^4$ \\
    $Q_{50}$ & $-3683.18 - 4855.97x + 2342.20x^2 + 509.59x^3 + 340.36x^4 + 324x^5$ \\
    $Q_{51}$ & $1960.80 + 2578.79x - 1263.64x^2 - 280.02x^3 - 167.77x^4 - 130.20x^5$ \\
    $Q_{52}$ & $-981.80 - 1286.88x + 643.53x^2 + 141.74x^3 + 72.23x^4 + 42.66x^5$ \\
    $Q_{53}$ & $463.12 + 604.69x - 309.13x^2 - 64.52x^3 - 25.87x^4 - 10.12x^5$ \\
    $Q_{54}$ & $-208.26 - 272.17x + 140.81x^2 + 25.87x^3 + 7.44x^4 + 1.33x^5$ \\
    $Q_{55}$ & $91.29 + 122.33x - 61.70x^2 - 9.53x^3 - 2.07x^4 - 0.04x^5$ \\
    \midrule
    \bottomrule
  \end{tabular}
\end{table*}

\section{Combined latent space projection and Volumetric Convolution with Roto-Translataional Kernels}
\label{app:convolution}

\noindent
\textbf{Theorem 1: } \textit{Suppose $f,g : X \longrightarrow \mathbb{R}^{3}$ are square integrable complex functions defined in $\mathbb{B}^{3}$ so that $\langle f,f \rangle < \infty$ and $\langle g,g \rangle < \infty$. Further, suppose $g$ is symmetric around north pole and $\tau (\alpha, \beta) = R_y(\alpha)R_z(\beta)$ where $R \in \mathbb{SO}(3)$ and $T_{r'}$ is translation of each point by $r'$. Then, }
\begin{equation}
\begin{split}
    f*g(r',\alpha,\beta) & \coloneqq  \int_{0}^1\int_{0}^{2\pi}\int_{0}^{\pi}P\{f(\theta, \phi, r)\}, T_{r'}\{\tau_{(\alpha, \beta)}(g(\theta, \phi, r))\}\sin\phi\, d\phi d\theta dr \\ 
    & \approx  \frac{4 \pi}{3} \sum\limits_{n=0}^{\infty} \sum\limits_{n'=0}^{\infty} \sum\limits_{l = 0}^{n} \sum\limits_{m = -l}^{l} \langle f_{nl}(r),  Q_{n'l}(r) \rangle ( e^{(n-l)r'}-  e^{(n'-l)r'}) \hat{\Omega}_{n,l,m}(f) \hat{\Omega}_{n',l,0} (g) Y_{l,m}(\alpha, \beta), 
\end{split}
\end{equation}
\textit{where $\hat{\Omega}_{n,l,m}(f), \hat{\Omega}_{n',l,0}(g)$ and $Y_{l,m}(\theta, \phi)$ are $(n,l,m)^{th}$ spectral moment of $f$, $(n',l,0)^{th}$ spectral moment of $g$, and spherical harmonics function respectively. $P\{\cdot\}$ is the projection to a latent space, $\tau (\alpha, \beta) = R_y(\alpha)R_z(\beta)$ where $R \in \mathbb{SO}(3)$ and $T_{r'}$ is translation of each point by $r'$.}

\noindent
\textbf{Proof}: The input function $f$ is projected to the latent space shape $\hat{f}$ by,

\begin{equation}
\label{equapp:reconstruction_latent}
\hat{f}(\theta, \phi, r) = \sum\limits_{n=0}^{\infty} \sum\limits_{l = 0}^{n} \sum\limits_{m = -l}^{l} \hat{\Omega}_{n,l,m}(f) \hat{Z}_{n,l,m}(\theta, \phi, r),
\end{equation}
where spectral moment $\hat{\Omega}_{n,l,m}(f)$ can be obtained using,
\begin{equation}
\label{omegaapp}
\hat{\Omega}_{n,l,m}(f) = \int_{0}^{1} \int_{0}^{2 \pi} \int_{0}^{\pi}  f(\theta, \phi, r) {\hat{Z}}^{\dagger}_{n,l,m} r^2 \sin\phi \, drd\phi d\theta.
\end{equation}
and,

\begin{equation}
    \hat{Z}_{n,l,m}(\theta, \phi, r) = \hat{Q}_{nl}(r)Y_{lm}(\theta, \phi),
\end{equation}
where,
\begin{equation}
\label{equ:w}
    \hat{Q}_{nl}(r) = f_{nl}(r) - \sum_{k=0}^{n-1} \sum_{m=0}^{k}W_{nlkm}\hat{Q}_{km}(r),
\end{equation}
\begin{equation}
    \label{equ:f}
    f_{nl} = (-1)^ln \sum_{k=0}^{n}\frac{((n-l)r)^k}{k!},
\end{equation}
and,
\begin{equation}
Y_{l,m} (\theta, \phi) = (-1)^m\sqrt{\frac{2l+1}{4\pi}\frac{(l-m)!}{(l+m)!}}P_l^m(\cos\phi)e^{im\theta},
\end{equation}
where $\phi \in [0,\pi]$ is the polar angle, $\theta \in [0, 2\pi ]$ is the azimuth angle, $l \in \mathbb{Z}^{+}$ is a non-negative integer, $m  \in \mathbb{Z}$ is an integer, $|m| < l$, and $P_l^m(\cdot)$ is the associated Legendre function,
\begin{equation}
P_l^m(x) = (-1)^m \frac{(1-x^2)^{m/2}}{2^ll!}\frac{d^{l+m}}{dx^{l+m}}(x^2-1)^l.
\end{equation}

In Eq.~\ref{equ:w}, the set $\{W_{nlkm}\}$ denotes trainable weights. Using this result, we can rewrite $f * g(r', \alpha, \beta)$ as,
\begin{equation}
\begin{split}
    f * g & (r, \alpha, \beta) =  \langle \sum\limits_{n=0}^{\infty} \sum\limits_{l = 0}^{n} \sum\limits_{m = -l}^{l} \hat{\Omega}_{n,l,m}(f) \hat{Z}_{n,l,m}(\theta, \phi, r), \\
    & T_{r'}\{\tau_{(\alpha, \beta)} (\sum\limits_{n'=0}^{\infty} \sum\limits_{l' = 0}^{n} \sum\limits_{m' = -l'}^{l'} \hat{\Omega}_{n',l',m'}(g) \hat{Z}_{n',l',m'}(\theta, \phi, r)))\} \rangle_{\mathbb{B}^3}
\end{split}
\end{equation}

Using the properties of inner product, this can be rewritten as,
\begin{equation}
\label{equation}
\begin{split}
    f * g  (r', \alpha, \beta) & = \sum\limits_{n=0}^{\infty} \sum\limits_{l = 0}^{n} \sum\limits_{m = -l}^{l} \sum\limits_{n'=0}^{\infty} \sum\limits_{l' = 0}^{'n} \sum\limits_{m' = -l'}^{l'} \hat{\Omega}_{n,l,m}(f)\hat{\Omega}_{n',l',m'}(g) \\
    &  \langle \hat{Z}_{n,l,m}(\theta, \phi, r), 
    T_{r'}\{\tau_{(\alpha, \beta)} (  \hat{Z}_{n',l',m'}(\theta, \phi, r))\} \rangle_{\mathbb{B}^3}
\end{split}
\end{equation}

Consider the term,
\begin{equation}
\begin{split}
    \Gamma & = \langle \hat{Z}_{n,l,m}(\theta, \phi, r), 
    T_{r'}\{\tau_{(\alpha, \beta)} (  \hat{Z}_{n',l',m'}(\theta, \phi, r)))\} \rangle_{\mathbb{B}^3} \\
    & =  \langle \hat{Q}_{nl}(r)Y_{lm}(\theta,\phi),  T_{r'}\{\tau_{(\alpha, \beta)} ( \hat{Q}_{n'l'}(r)Y_{l'm'}(\theta,\phi))\} \rangle_{\mathbb{B}^3}
\end{split}
\end{equation}

$\Gamma$ can be decomposed into its angular and linear components as,
\begin{equation}
\label{equ:gamma}
\begin{split}
    \Gamma = & \int_0^1 \hat{Q}_{nl}(r)T_{r'}\{\hat{Q}_{n'l'}(r)\} r^2dr  \int_{0}^{2\pi}\int_{0}^{\pi} Y_{lm}(\theta,\phi)\tau_{(\alpha, \beta)} (Y_{l'm'}(\theta,\phi))\}sin\phi d\phi d\theta.
\end{split}
\end{equation}

First, consider the angular component, 
\begin{equation}
\label{equ:ang}
    Ang(\Gamma) = \int_{0}^{2\pi}\int_{0}^{\pi} Y_{lm}(\theta,\phi)\tau_{(\alpha, \beta)} (Y_{l'm'}(\theta,\phi))\}sin\phi d\phi d\theta.
\end{equation}

Since $g(\theta, \phi, r)$ is symmetric around $y$, using the properties of spherical harmonics, Eq.~\ref{equ:ang} can be rewritten as,
\begin{equation}
\begin{split}
        Ang(\Gamma) =  \int_{0}^{2\pi}\int_{0}^{\pi}& Y_{lm}(\theta,\phi)  \sum_{m''=-l'}^{l'} Y_{l',m''}D^{l'}_{m''0}(\alpha, \beta)\}sin\phi d\phi d\theta
\end{split}
\end{equation}
where $D^{l'}_{mm'}$ is the Wigner-D matrix. But $D^{l'}_{m''0} = Y_{l',m''}$, and hence,
\begin{equation}
\begin{split}
        Ang(\Gamma) =  \sum_{m''=-l'}^{l'} & Y_{l',m''}(\alpha, \beta)  \int_{0}^{2\pi}\int_{0}^{\pi} Y_{lm}(\theta,\phi) Y_{l',m''}(\theta,\phi)\}sin\phi d\phi d\theta
\end{split}
\end{equation}
Since spherical harmonics are orthogonal,
\begin{equation}
    \label{equ:angresult}
    Ang(\Gamma) =  C_{ang}Y_{l,m}(\alpha, \beta), 
\end{equation}
where $C_{ang}$ is a constant. Consider the linear component of Eq.~\ref{equ:gamma}. It is important to note that for simplicity, we derive equations for the orthogonal case and use the same results for non-orthogonal case. In practice, this step does not reduce accuracy.

\begin{equation}
\label{equ:lin}
    Lin(\Gamma) = \int_0^1 \hat{Q}_{nl}(r)T_{r'}\{\hat{Q}_{n'l'}(r)\} r^2dr.
\end{equation}
Since $\hat{Q}_{nl}(r)$ is a linear combination of $f_{nl}  \approx  (-1)^ln\exp(r(n-l))$, it is straightforward to see that,
\begin{equation}
\label{equ:translation}
\begin{split}
    Q_{nl}(r + r') = & f_{nl}(r)\exp((n-l)r')  - \sum_{k=0}^{n-1} \sum_{m=0}^{k}C_{nlkm}Q_{km}(r)\exp(k-m)r'.
\end{split}
\end{equation}
Also, we have derived that $l=l'$ from the result in Eq.~\ref{equ:angresult}. Applying this result and Eq.~\ref{equ:translation} to Eq.~\ref{equ:lin} gives,
\begin{equation}
\begin{split}
         \langle Q_{nl}(r+r'), & Q_{n'l}(r) \rangle 
          = \langle f_{nl}(r+r'),  Q_{n'l}(r)\rangle - \sum_{k=0}^{n-1} \sum_{m=0}^{k}C_{nlkm} \langle Q_{km}(r+r'), Q_{n'l}(r) \rangle,
\end{split}
\end{equation}
\begin{equation}
\begin{split}
         \langle Q_{nl}(r+r'), & Q_{n'l}(r) \rangle  = \langle f_{nl}(r),  Q_{n'l}(r) \rangle e^{(n-l)r'}  - \sum_{k=0}^{n-1} \sum_{m=0}^{k}C_{nlkm}\langle Q_{km}(r)e^{(k-m)r'}, Q_{(n'l}(r) \rangle.
\end{split}
\end{equation}

Since $Q_{km}$ and $Q_{n'l}$ are orthogonal,
\begin{equation}
\begin{split}
        \langle Q_{nl}(r+r'), & Q_{n'l}(r) \rangle  = \langle f_{nl}(r),  Q_{n'l}(r) \rangle e^{(n-l)r'}   - C_{nln'l} e^{(n'-l)r'} ||Q_{n'l}||^2.
\end{split}
\end{equation}
But since for orthogonal case, $C_{nln'l'} = \frac{<f_{nl}, Q_{n'l'}>}{||Q_{n'l'}||^2}$,
\begin{equation}
\begin{split}
         \langle Q_{nl}(r+r'),& Q_{n'l}(r) \rangle = \langle f_{nl}(r),  Q_{n'l}(r) \rangle e^{(n-l)r'}   - \langle f_{nl}(r), Q_{n'l'}(r) \rangle e^{(n'-l)r'},
\end{split}
\end{equation}
\begin{equation}
\label{equ:linresult}
     \langle Q_{nl}(r+r'), Q_{n'l}(r) \rangle = \langle f_{nl}(r),  Q_{n'l}(r)\rangle ( e^{(n-l)r'}-  e^{(n'-l)r'}).
\end{equation}
Combining Eq.~\ref{equ:angresult} and Eq.~\ref{equ:linresult} for Eq.~\ref{equation} and choosing the normalization constant to be $\frac{4\pi}{3}$ (since the integration is over unit ball) gives, 
\begin{equation}
\begin{split}
 & f*g(r',\alpha,\beta) \approx   \frac{4 \pi}{3} \sum\limits_{n=0}^{\infty} \sum\limits_{n'=0}^{\infty} \sum\limits_{l = 0}^{n} \sum\limits_{m = -l}^{l} \langle f_{nl}(r),  Q_{n'l}(r) \rangle ( e^{(n-l)r'}-  e^{(n'-l)r'}) \hat{\Omega}_{n,l,m}(f) \hat{\Omega}_{n',l,0} (g) Y_{l,m}(\alpha, \beta).
\end{split}
\end{equation}
Q.E.D.

\section{Ablation study on input point cloud density}
A critical problem associated with directly consuming point clouds, in order to learn features, is the redundancy of information. This property hampers optimal feature learning using neural network based models, by imposing an additional overhead. To verify this, we conduct an ablation study on the density of the input point cloud, and observe the performance variations of our model. The obtained results are reported in Table \ref{tab:density}. As the results suggest, there is no clear variation of classification performance, although the input sampling density is increased. Therefore, it can be empirically concluded that input point clouds are not optimal to be directly fed to learning networks, due to their inherent redundancy. As a result, significant reduction in their density could still lead to comparable performance with that of the original point cloud. 

 \begin{table}
    \caption {Ablation study on the input point cloud density. We sample the input points on a grid (r= 25, $\theta$ = 36, $\phi$ = 18) before feeding to the network.}
\centering
\label{tab:density}
\scalebox{0.9}{
 \begin{tabular}{  c  c   }
     \hline
 \textbf{Original point cloud sampling} & \textbf{Accuracy}  \\ 
\hline
(r= 250, $\theta$ = 200, $\phi$ = 200) & 94.22\% \\ 
(r= 300, $\theta$ = 250, $\phi$ = 250) & 94.21\%\\
(r= 400, $\theta$ = 300, $\phi$ = 300) &  \textbf{94.23\%}\\
(r= 500, $\theta$ = 400, $\phi$ = 400) &  94.20\%\\
\hline
  \end{tabular}}
  \end{table}

\end{document}